%% file: CameraReady2027.tex
\documentclass[letterpaper]{article} 
\usepackage[preprint]{aaai2027}  
\usepackage[hyphens]{url}  
\usepackage{graphicx} 
\usepackage{natbib}  
\usepackage{caption} 
\usepackage{algorithm}
\usepackage{algorithmic}
\usepackage{amssymb}
\usepackage{amsmath}
\usepackage{multirow}
\usepackage[table]{xcolor}

\usepackage{newfloat}
\usepackage{listings}
\DeclareCaptionStyle{ruled}{labelfont=normalfont,labelsep=colon,strut=off} 
\floatstyle{ruled}
\newfloat{listing}{tb}{lst}{}
\floatname{listing}{Listing}

\usepackage{booktabs}

\title{\textbf{\textit{StrataVLA}}: Hierarchical and Efficient 3D Geometric Grounding for Vision-Language-Action Models}
\author {
    Jin Cui\textsuperscript{\rm 1}\equalcontrib,
    Zhaoyu Pu\textsuperscript{\rm 2}\equalcontrib,
    Botao Cai\textsuperscript{\rm 3},
    Jun Ye\textsuperscript{\rm 1,\rm 3},
    Xinyue Long\textsuperscript{\rm 1,\rm 3},
    Boran Zhao\textsuperscript{\rm 1,\rm 3}\corresponding,
    Pengju Ren\textsuperscript{\rm 1}
}
\affiliations {
    \textsuperscript{\rm 1}State Key Laboratory of Human-Machine Hybrid Augmented Intelligence,\\
    and Institute of Artificial Intelligence and Robotics, Xi'an Jiaotong University\\
    \textsuperscript{\rm 2}School of Computer Science and Technology, Xi'an Jiaotong University \\
    \textsuperscript{\rm 3}School of Software Engineering, Xi'an Jiaotong University \\
    andycui@stu.xjtu.edu.cn
}

\begin{document}

\maketitle

\begin{abstract}
Vision-Language-Action (VLA) models inherit strong semantic priors
from large-scale vision-language pretraining, yet remain limited in
robotic manipulation by insufficient 3D spatial awareness. Existing
approaches either require explicit depth or point-cloud inputs,
compress geometry into training-time supervision, or inject it only
at the model input or action expert, leaving the vision-language
backbone without persistent access to task-relevant spatial information. We introduce \textbf{\textit{StrataVLA}}, a plug-and-play framework for hierarchical geometric grounding. A frozen geometry foundation model extracts shared geometric features from RGB observations, while sparse, layer-specific \textbf{\textit{Geometry Adapters}} allow visual representations at selected backbone depths to retrieve relevant geometric evidence through cross-attention. To make inference-time geometry practical, \textbf{\textit{StrataVLA}} further combines task-aware routing with an LRU feature cache that exploits temporal redundancy during task manipulation. Experiments on \textsc{LIBERO}, SimplerEnv, and real-world manipulation demonstrate consistent gains over strong VLA baselines. \textbf{\textit{StrataVLA}} achieves 98.53\% average success on LIBERO suites while reducing geometry-model invocations by up to 88\%, establishing hierarchical geometry injection as an effective and efficient way to spatially grounded robotic control.
\end{abstract}


\section{Introduction}

Recent advances in Vision-Language Models (VLMs) have unlocked a promising pathway toward generalizable robotic manipulation. As end-to-end learning paradigms have matured, the field has progressively shifted toward Vision-Language-Action (VLA) models—a unified framework in which a single model ingests visual observations and natural language instructions, leverages rich semantic priors acquired from internet-scale pretraining, and directly outputs continuous action trajectories \citep{Brohan2023RT2VM, Kim2024OpenVLAAO, Black20240AV}. Within this paradigm, VLAs have demonstrated impressive task adaptability: they perform scene understanding grounded in semantic priors and translate high-level language intent into manipulation behavior.

Despite this progress, VLA performance remains fundamentally limited by \textbf{\textit{reliable 3D spatial perception}}. VLM backbones are trained predominantly on 2D image-text data and therefore excel at recognizing \emph{what} is present, whereas manipulation additionally requires understanding \emph{where} objects lie in 3D and \emph{how} they relate geometrically. Projection ambiguity becomes particularly severe under occlusion, depth-dependent object configurations, and precision manipulation. Consequently, current VLAs often fail precisely in scenarios where accurate geometry, depth, and spatial relations are essential \citep{Li2025SpatialFI, Guo2025GLaDGL}.

Existing attempts to introduce 3D awareness into VLAs follow four main directions. (1) \textit{Explicit 3D methods} augment policy observations with depth maps or point clouds \citep{Li2025PointVLAIT, Qu2025SpatialVLAES, Sun2025GeoVLAE3}. Although effective, they incur nontrivial 3D processing costs and are difficult to scale to predominantly RGB-only robot datasets.  
(2) \textit{Distillation-based methods} align VLA visual representations with features from frozen 3D foundation models during training \citep{Li2025SpatialFI, Guo2025GLaDGL}, achieving efficient inference but reducing geometry to a static supervision signal; after the teacher is removed, the policy must reconstruct fragile task-specific geometry from its 2D representations. 
(3) \textit{Inference-time fusion methods} preserve 3D information by replacing the 2D encoder with a geometric backbone \citep{Abouzeid2025GeoAwareVLAIG} or injecting geometric features only at the VLM input \citep{Lin2025Evo0VM, Yu20263DMixFV}, yet they either sacrifice the rich semantic priors of image-text pretraining or provide only a shallow, one-shot geometric signal that deeper reasoning layers cannot adaptively query. 
Finally, (4) \textit{action-expert injection methods} feed 3D features directly into the diffusion or flow-matching controller \citep{Sun2025GeoVLAE3, Rao2026AugVLA3DDF}, improving low-level trajectory precision but leaving the VLM backbone—the module responsible for language grounding, object selection, and semantic scene reasoning—geometrically under-informed. Taken together, these observations converge on a clear question: 
\textit{How can task-relevant geometry remain accessible throughout a VLA backbone while preserving pretrained representations and maintaining practical inference efficiency?}

To this end, we propose \textbf{\textit{StrataVLA}}, which empowers VLA models with hierarchical 3D geometric grounding for robot manipulation. \textbf{\textit{StrataVLA}} bridges the frozen VGGT geometry foundation model~\citep{Wang2025VGGTVG} with the VLM backbone through a plug-and-play \textbf{\textit{Geometry Adapter (GA)}}. Rather than injecting 3D features once at the input, each of the selected layers of the LLM backbone hosts an independent \textbf{\textit{GA}} module that performs multi-head cross-attention between the layer's current image token representations and the shared VGGT geometric features, enabling every layer to actively extract the most geometrically relevant information for its current representational state. 

We further address the computational cost of inference-time geometry. We therefore introduce a task-aware gating mechanism that predicts a continuous injection strength from task instruction semantics and visual scene complexity, routing computation on demand. A 3D feature LRU cache further reduces overhead by exploiting the temporal redundancy of fixed-camera robot scenes. These mechanisms reduce total VGGT invocation frequency by up to 88\% without compromising performance. Extensive experiments on LIBERO, SimplerEnv, and two in-house real-world manipulation suites demonstrate consistent improvements over strong generalist and spatially enhanced baselines. The three sparse GAs introduce only 3.8\% latency overhead when geometry is available, while the complete routing-and-cache system limits the amortized end-to-end overhead to 23.0\%.Our main contributions are threefold:
\begin{itemize}
    \item We propose \textbf{\textit{StrataVLA}}, a plug-and-play
    framework for hierarchical, inference-time persistent, and
    training-stable geometric grounding of VLA backbones.

    \item We introduce sparse, layer-specific
    \textbf{\textit{Geometry Adapters}} with zero-initialized residual
    injection, enabling visual representations at different depths
    to retrieve task-relevant evidence from shared geometric features.

    \item We introduce task-aware routing with feature reuse to make inference-time geometry efficient, and show
    consistent improvements across simulation and real-world manipulation with modest amortized latency overhead.
\end{itemize}


\section{Related Work}
\paragraph{3D-Enhanced Vision-Language-Action Models.}
Existing 3D-enhanced VLAs mainly differ in how geometry enters the policy. Explicit approaches introduce depth maps, point clouds, or structured 3D representations \citep{Li2025PointVLAIT, Qu2025SpatialVLAES, Bhat20253DCL, Sun2025GeoVLAE3}, improving spatial precision but requiring additional sensing or costly 3D processing. Distillation methods transfer geometric knowledge from frozen 3D models into visual tokens during training \citep{Li2025SpatialFI, Guo2025GLaDGL, Sun2026ROCKETRM}, yielding efficient inference but removing direct access to geometry at deployment. Other methods retain geometry at inference by replacing the visual encoder \citep{Abouzeid2025GeoAwareVLAIG}, fusing geometric features only at the VLM input \citep{Lin2025Evo0VM, Yu20263DMixFV}, or injecting them into the action expert \citep{Li2025PointVLAIT, Sun2025GeoVLAE3, Rao2026AugVLA3DDF, Yang2026ABotM0VF}. These designs either weaken pretrained semantic representations or restrict geometry to a single stage. In contrast, \textbf{\textit{StrataVLA}} preserves the original visual backbone and enables selected layers to repeatedly retrieve task-relevant evidence from a shared geometric representation.

\noindent{\textbf{Hierarchical and Stable Backbone Adaptation.}}
VLM representations exhibit clear depth-wise specialization: shallow layers preserve local visual structure, middle layers support cross-modal grounding, and deeper layers consolidate task-level semantics \citep{Kaduri2024WhatsIT, Jiang2024DevilsIM}. Visual grounding may nevertheless decay toward the output \citep{Li2025TheHL}, motivating geometric injection at multiple backbone depths rather than a single input-level fusion; related evidence further shows that distributed layer selection outperforms late-only adaptation \citep{Abouzeid2025GeoAwareVLAIG}. Such modification must also preserve pretrained capabilities, since task-specific fine-tuning can cause catastrophic forgetting \citep{Goodfellow2013AnEI, Biderman2024LoRALL}. \textbf{\textit{StrataVLA}} connects each Geometry Adapter through a zero-initialized residual projection, allowing geometric grounding to emerge gradually without perturbing the initial policy.

\section{Method}
\label{sec:method}

\begin{figure*}
    \centering
    \includegraphics[width=1\linewidth]{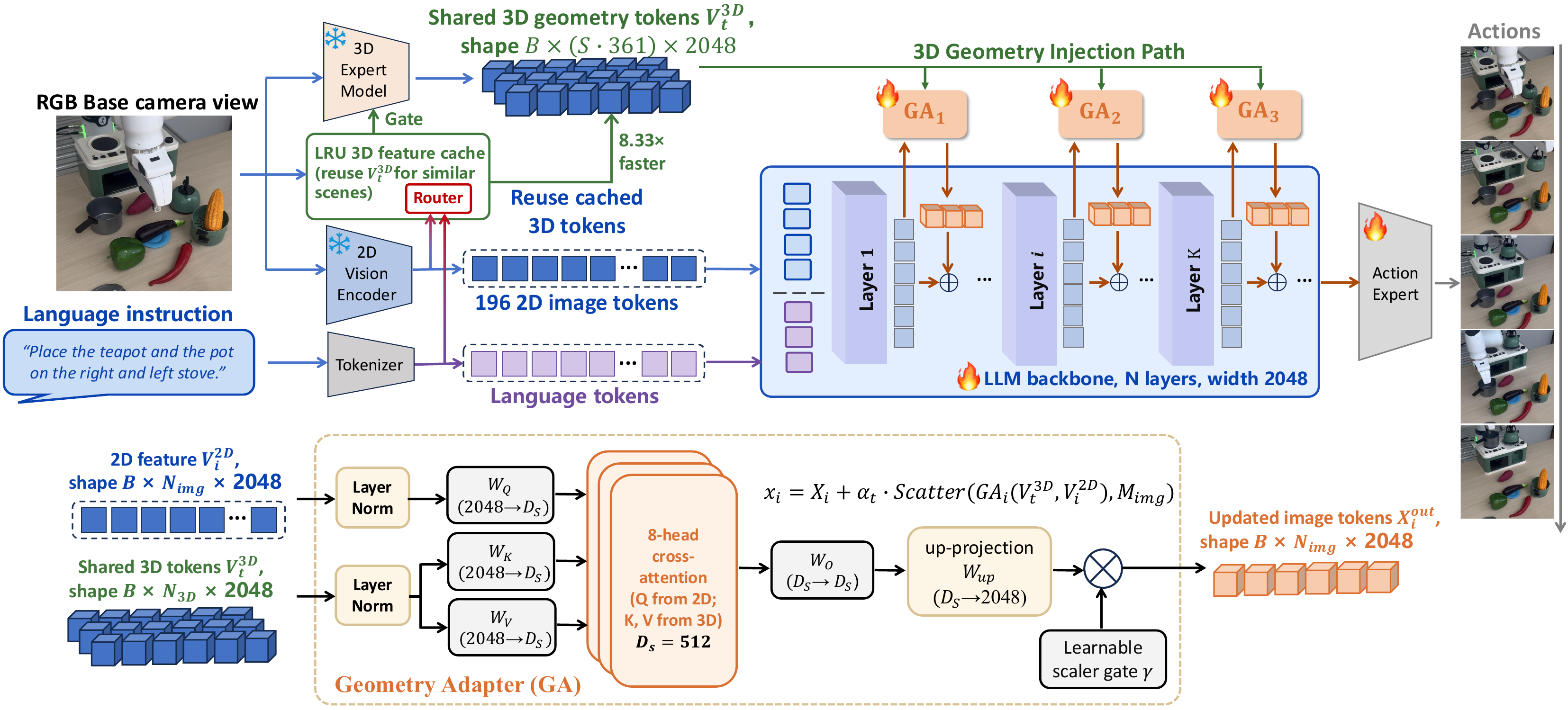}
     \vspace{-3mm}
    \caption{Overview of StrataVLA. A frozen visual geometry foundation model provides a shared geometric memory $V^{3d}$, while the 2D vision encoder and tokenizer produce visual and language inputs for the VLM backbone. At selected backbone layers, independent \textbf{\textit{Geometry Adapters (GAs)}} use the current image tokens as queries and the shared geometry tokens as keys and values to retrieve layer-relevant geometric evidence. A task-aware geometry router and an LRU feature cache bypass or reuse geometry computation when possible, and the resulting geometry-enhanced prefix conditions the action generation process.}
    \label{fig:placeholder}
    \vspace{-2mm}
\end{figure*}

\subsection{Overview}
Given RGB observations $\mathcal{I}_t=\{I_t^{s}\}_{s=1}^{S}$, a language instruction $\ell$, and the robot state $p_t$, model predicts an action chunk $a_t\in\mathbb{R}^{H_a\times D_a}$. \textbf{\textit{StrataVLA}} builds on
$\pi_0$~\cite{Black20240AV}, which contains a PaliGemma vision--language backbone and a lightweight Action Expert coupled through layer-wise joint attention. Vision encoder encodes the input images into $N_{\rm img}$ visual tokens
$V^{2d}_0\in\mathbb{R}^{B\times N_{\rm img}\times D}$, while the instruction is embedded as language tokens $T_0$. Their concatenation forms the prefix $x^p_0=[V^{2d}_0;T_0]$; the proprioceptive state, action chunk, and flow time form the action suffix $x^a_0$. A frozen 3D expert produces a shared geometric memory $V^{3d}_t$. Each of the $n=18$ backbone layers contains an independent \textbf{\textit{Geometry Adapter (GA)}}, allowing its current image tokens to query this memory before being passed to the next layer. A task-aware gate controls both the residual injection strength and whether the geometry branch is
executed.

\subsection{Geometric Memory Extraction}
We resize the selected camera views to $518\times518$ and feed them to the frozen VGGT aggregator. Let $\mathcal{A}^{(24)}(\mathcal{I}_t)$ denote its final-layer output. VGGT produces one camera token, four register tokens, and a $37\times37$ patch grid per view, with a $2048$-dimensional concatenated frame/global representation. We discard the five special tokens, reshape the remaining patches spatially, and apply stride-$2$
subsampling:
\begin{equation}
\begin{aligned}
G_t &=
\operatorname{reshape}\!\left(
\mathcal{A}^{(24)}(\mathcal{I}_t)[:,:,5:,:]\right)
\in\mathbb{R}^{B\times S\times37\times37\times2048},\\
V^{3d}_t &=
\operatorname{Flatten}\!\left(G_t[:,:,::2,::2,:]\right),\\
V^{3d}_t &\in
\mathbb{R}^{B\times(361S)\times D_g},
\qquad D_g=2048.
\end{aligned}
\label{eq:vggt_feature}
\end{equation}
This reduces the geometric sequence by $73.6\%$ while preserving its
spatial layout. Because the cameras are fixed in our manipulation setting,
we do not inject explicit calibration tokens. VGGT is evaluated under FP16
autocasting and its output is detached, making it a parameter- and
activation-free feature provider at optimization time.

\subsection{Hierarchical Geometric Injection}
\label{sec:hierarchical_injection}

The geometric memory extracted above provides a unified description of scene structure, but directly fusing it with the input visual tokens yields only a one-shot geometric signal. As representations evolve across the backbone, different layers may require distinct geometric evidence: early layers primarily refine local appearance and spatial boundaries, whereas deeper layers progressively associate geometry with task semantics and action-relevant entities. We therefore introduce a layer-wise \textbf{\textit{Geometry Adapter (GA)}}, allowing the visual representations at each depth to independently query the shared geometric memory.

Let $\mathcal{F}_i$ denote the $i$-th backbone block and
$X_i\in\mathbb{R}^{B\times N\times D}$ its output:
\begin{equation}
X_i=\mathcal{F}_i(x_{i-1}), \qquad i=1,\ldots,L.
\end{equation}
A fixed visual-token mask $M_{\rm img}$ extracts the image states $V_i^{2d}=\operatorname{Gather}(X_i,M_{\rm img})$, while all language, state, and action-related tokens remain outside the injection path. Given the shared geometric memory
$V^{3d}\in\mathbb{R}^{B\times N_g\times D_g}$, the $i$-th \textbf{\textit{GA}} first normalizes both modalities and maps them into a common interaction space of width $D_s$:
\begin{equation}
\begin{aligned}
Q_i &= \operatorname{LN}_{2d}(V_i^{2d})W_i^Q,\\
K_i &= \operatorname{LN}_{3d}(V^{3d})W_i^K,\\
U_i &= \operatorname{LN}_{3d}(V^{3d})W_i^V .
\end{aligned}
\label{eq:cvge_qkv}
\end{equation}
The current visual states act as queries, whereas the geometric tokens serve as keys and values. Multi-head cross-attention therefore retrieves geometry according to the layer's current representational context:
\begin{equation}
Z_i=
\operatorname{Concat}_{h=1}^{H}
\left[
\operatorname{softmax}
\left(
\frac{Q_i^h(K_i^h)^\top}{\sqrt{d_h}}
\right)U_i^h
\right]W_i^O .
\label{eq:cvge_attention}
\end{equation}
The attended features are projected back to the backbone width and modulated by a learnable layer-specific scalar:
\begin{equation}
\Delta_i=
\gamma_i\left(Z_iW_i^{\rm up}+b_i^{\rm up}\right).
\label{eq:cvge_output}
\end{equation}
Although all \textbf{\textit{GAs}} share the same geometric memory, they use independent parameters, enabling each depth to learn its own correspondence between visual semantics and 3D structure.

Finally, the geometry-enhanced features are scattered back only to the
visual-token positions through a gated residual connection:
\begin{equation}
x_i =
X_i+
\alpha_t\,
\operatorname{Scatter}(\Delta_i,M_{\rm img}),
\label{eq:hierarchical_residual}
\end{equation}
where $\alpha_t$ is the task-aware routing coefficient introduced in the next subsection. We initialize $W_i^{\rm up}$ and $b_i^{\rm up}$ to zero and set $\gamma_i$ to a nonzero value. Consequently, the added branch initially produces exactly zero output, preserving the behavior of the pretrained backbone while allowing geometric information to emerge gradually during optimization.

\subsection{Task-Aware Geometry Routing and Reuse}
Executing VGGT at every control step is unnecessary for visually simple tasks. We therefore predict a frame-level injection strength from instruction semantics and scene complexity. A text head pools the initial language tokens, while a visual head pools the visual tokens and predicts occlusion and clutter scores:
\begin{equation}
\begin{aligned}
s_t^{\rm text}
&=\sigma\!\left(
g_{\rm text}(\operatorname{MeanPool}(T_0))\right),\\
(p_t^{\rm occ},p_t^{\rm crowd})
&=\sigma\!\left(
g_{\rm vis}(\operatorname{MeanPool}(V^{2d}_0))\right),\\
s_t^{\rm vis}
&=\max(p_t^{\rm occ},p_t^{\rm crowd}).
\end{aligned}
\end{equation}
Both heads use a $2048\!\rightarrow\!256$ GELU projection; the visual heads share the projection layer. To avoid injecting unreliable geometry from blurred or severely exposed frames, we additionally compute a parameter-free quality score
\begin{equation}
\begin{aligned} \small
q_t={}&0.6\,\operatorname{clip}\!\left(
\frac{\operatorname{Var}_{HW}(\bar I_t)}{0.1},0,1\right)\\
&+0.4\,\operatorname{clip}\!\left(
1-r_t^{>0.95}-r_t^{<0.05},0,1\right)
\end{aligned}
\end{equation}
where $r_t^{>0.95}$ and $r_t^{<0.05}$ are the fractions of over- and under-exposed pixels. The final gate is
\begin{equation}
\alpha_t
=\mathbf{1}[q_t\geq\tau_q]\,q_t
\cdot\sigma\!\left(
w_{\rm text}s_t^{\rm text}
+w_{\rm vis}s_t^{\rm vis}+b\right).
\label{eq:task_gate}
\end{equation}
We set $\tau_q=0.25$ and initialize $w_{\rm text}=w_{\rm vis}=1.5$ and $b=1$, biasing early training toward using geometry. At inference, values below a small routing threshold are treated as zero; the model then bypasses both VGGT and all \textbf{\textit{GAs}} and reduces exactly to the base model.

For active routes, we exploit the temporal redundancy of fixed-camera manipulation using a GPU-resident LRU cache. An $8\times8$ differential perceptual hash is computed for every camera and concatenated into a joint key $h_t$. The geometric memory is obtained as
\begin{equation}
\widetilde V_t^{3d}=
\begin{cases}
\mathcal{C}[h_t], & h_t\in\mathcal{C},\\
E_{\rm VGGT}(\mathcal{I}_t), & \text{otherwise},
\end{cases}
\end{equation}
where a miss is inserted into a capacity-$8$ LRU cache after detaching and cloning the tensor. The cache is cleared at episode boundaries and disabled during training.

\subsection{Flow-Matching Action Generation and Inference}
\textbf{\textit{StrataVLA}} adopts the original action formulation. For expert action chunk $a$, Gaussian noise $\epsilon\sim\mathcal{N}(0,I)$, and flow time $\tau\sim\mathcal{U}(0,1)$, we construct
\begin{equation}
x_\tau=(1-\tau)a+\tau\epsilon,
\qquad
u_\tau=\epsilon-a,
\end{equation}
and train the Action Expert to predict the conditional velocity $v_\theta(x_\tau,\tau\mid\mathcal{I}_t,\ell,p_t)$. During inference, the geometry-enhanced prefix is evaluated once and its per-layer keys and values are cached. The action suffix is then integrated from noise to an action chunk using ten Euler denoising steps; neither VGGT nor \textbf{\textit{GA}} is recomputed inside this loop because their information is already encoded in the prefix KV cache.

\paragraph{Training strategy.}
VGGT remains frozen throughout training. We jointly optimize SigLIP, the PaliGemma backbone, the Action Expert and output projection, all \textbf{\textit{GA}} modules, and the dual-modal routing heads. For the first updates, we force the geometry route on ($\alpha_t=1$) to warm up the zero-initialized
adapters; afterward, Eq.~\eqref{eq:task_gate} is used end-to-end. The primary objective is the flow-matching regression loss
$\mathcal{L}_{\rm flow}=
\mathbb{E}_{\tau,\epsilon}
\|v_\theta(x_\tau,\tau)-u_\tau\|_2^2$.
When a task-level geometry label $y_{\rm gate}$ is available, we add
$\mathcal{L}_{\rm gate}=
\operatorname{BCE}(\alpha_t,y_{\rm gate})$, using weak labels from
spatial-relation instructions and offline task-level gain estimates;
unlabeled samples contribute only $\mathcal{L}_{\rm flow}$. The complete objective is $\mathcal{L}=\mathcal{L}_{\rm flow}+0.05\mathcal{L}_{\rm gate}$. PaliGemma and the Action Expert run in BF16, VGGT in FP16, and \textbf{\textit{GA}} LayerNorms in FP32; gradient checkpointing is used for the 18 joint-attention layers, while the inference-only geometry cache is disabled.

\input{Tables/main-exp}

\section{Experiments}
\label{sec:experiments}

\subsection{Experimental Setup}
\label{sec:experimental_setup}

\paragraph{Evaluation.}
We evaluate \textit{\textbf{StrataVLA}} on two public simulation benchmarks, LIBERO~\cite{Liu2023LIBEROBK} and SimplerEnv~\cite{Li2024EvaluatingRR}, and an
in-house real-world manipulation suite. 

\textbf{\textit{LIBERO.}}
We evaluate on the four standard LIBERO suites:
LIBERO-Spatial, LIBERO-Object, LIBERO-Goal, and LIBERO-Long. Each suite contains ten tasks with 50 expert demonstrations per task and evaluates generalization over spatial arrangements, object identities, task goals, and long-horizon behavior, respectively. 

\textbf{\textit{SimplerEnv.}} We evaluate real-to-sim generalization on the Google Robot suite, including \emph{Pick Coke Can}, \emph{Move Near}, and \emph{Open/Close Drawer}, following the standard visual-matching protocol. We do not include the WidowX suite since a faithful evaluation in this setting requires additional BridgeData-specific observation, action-space, and embodiment adaptation for the $\pi_0$ checkpoint, and its principal manipulation skills are already included in our real-world evaluation.

\paragraph{Model and implementation details.}
We instantiate \textit{\textbf{StrataVLA}} on two representative VLA families:
$\pi_0$~\cite{Black20240AV} and
OpenVLA-OFT~\cite{Kim2025FineTuningVM}. The $\pi_0$ policy contains a 3B PaliGemma model, composed of a
SigLIP-So400m visual encoder and a Gemma-2B language backbone, together with a 300M-parameter flow-matching Action Expert.
OpenVLA-OFT is built from the 7B OpenVLA policy, which
fine-tunes a Prismatic VLM comprising a fused SigLIP--DINOv2 visual
encoder, a visual projector, and a Llama-2 7B decoder. We retain the
OFT formulation with parallel action decoding, action chunking,
continuous actions, and its native $\ell_1$ regression objective.

We insert three independent \textbf{\textit{GA}} modules at shallow, middle, and deep decoder depths. Using zero-based indexing, they are attached after layers $\{2,8,13\}$ of the 18-layer PaliGemma backbone and after layers $\{4,15,24\}$ of the 32-layer OpenVLA backbone; both positions are selected by our experiment detailed in the following section. The \textbf{\textit{GAs}} and task-aware router add approximately 15M trainable parameters to the $\pi_0$ variant and 22M to the OpenVLA-OFT variant, corresponding to less than $0.5\%$ of either base policy. The geometry expert remains frozen.

For LIBERO, we train our policy on the mixture of the four standard suites for 80K gradient steps. Specifically, we first optimize only the GAs and routing heads for 8K steps with the pretrained policy frozen, and then jointly fine-tune the complete policy for 72K steps while keeping VGGT frozen. SimplerEnv follows the same training setting. For the real-world experiments, we collect 80 teleoperated demonstrations for each of five tasks, yielding 400 episodes, with an average trajectory length of 280 steps. We train our policy for 30K gradient updates in total with 3K:27K division.

Policy images are resized to $224\times224$, while the same frames are independently resized to the 3D expert's input resolution. We preserve the native action objective of each base model: flow matching for $\pi_0$ and continuous $\ell_1$ regression for OpenVLA-OFT. Training is distributed over 8 NVIDIA RTX 6000 Ada GPUs. For real-world deployment, the policy receives RGB observations from
one third-person Intel RealSense D435i camera and is embodied on FrankaResearch 3 robot. Policies are executed on an NVIDIA RTX 4090 GPU.

\paragraph{Baselines.}
On LIBERO, we compare against representative generalist VLAs and
spatially enhanced manipulation policies listed in Table.~\ref{tab:main_results}. On SimplerEnv, we additionally
compare with RT-1-X, RT-2-X~\cite{brohan2023rt2visionlanguageactionmodelstransfer}, RoboVLM~\cite{Li2024WhatMI}, and FALCON~\cite{Zhang2025FromST}.
We report published results obtained under matching benchmark protocols whenever available.

\subsection{Main results}

 As shown in Table~\ref{tab:main_results}, \textit{\textbf{StrataVLA}} achieves state-of-the-art performance across both simulation and real-world evaluations. On LIBERO-Spatial, where accurate spatial reasoning is most critical, our model attains a perfect success rate of $100.0\%$, while also achieving the best results on LIBERO-Object and LIBERO-Goal, yielding an average success rate of $98.53\%$ across the four suites. Consistent gains are observed on the SimplerEnv Google Robot tasks, including improvements in object grasping, relative placement, and articulated-object manipulation. These results indicate that injecting geometric representations effectively complements the semantic knowledge acquired through image--text pretraining and embodied post-training. The advantage becomes more pronounced in the real world: the Kitchen tasks require identifying task-relevant objects under severe clutter and recovering sufficiently accurate geometry for grasping, whereas the Room tasks additionally demand reasoning over relative spatial relations, contact-point selection, and precise end-effector displacement. Existing VLAs, including the strong $\pi_{0.5}$ baseline, remain unreliable under these conditions, suggesting that 2D-centric embodied training alone does not provide sufficient 3D spatial awareness. In contrast to encoder replacement, shallow input fusion, or geometry injection to the action expert, \textbf{\textit{GA}} preserves the pretrained semantic representation while enabling visual tokens at multiple backbone depths to retrieve task-relevant 3D evidence, resulting in more effective and robust spatial grounding.

 \paragraph{Injection Layer Selection.}
Figure~\ref{fig:injection} analyzes where geometric features should be introduced into the backbone. Injecting \textbf{\textit{GA}} into every layer provides no consistent advantage over sparse three-layer injection: the two settings achieve comparable performance on LIBERO and Google Robot, while the sparse variant performs slightly better in the real-world evaluation and exhibits more stable optimization. This behavior is consistent with the functional specialization of VLMs across depth, where shallow, intermediate, and deep layers respectively emphasize local visual structure, cross-modal grounding, and task-level semantics. The learned injection ratios further reveal a clear hierarchical pattern, with distinct response peaks appearing within each depth range across LIBERO-Spatial, Object, and Goal. We therefore select one representative layer from each stage rather than densely modifying the entire backbone. Additional experiments show that performance is largely insensitive to the exact layer chosen within the corresponding shallow, middle, or deep range, providing substantial flexibility when adapting \textbf{\textit{GA}} to different VLA architectures. Notably, LIBERO-Spatial consistently produces the strongest injection magnitude, indicating that \textbf{\textit{GA}} automatically allocates more geometric information to tasks with greater spatial-reasoning demands.

\begin{figure*}
    \centering
    \includegraphics[width=1\linewidth]{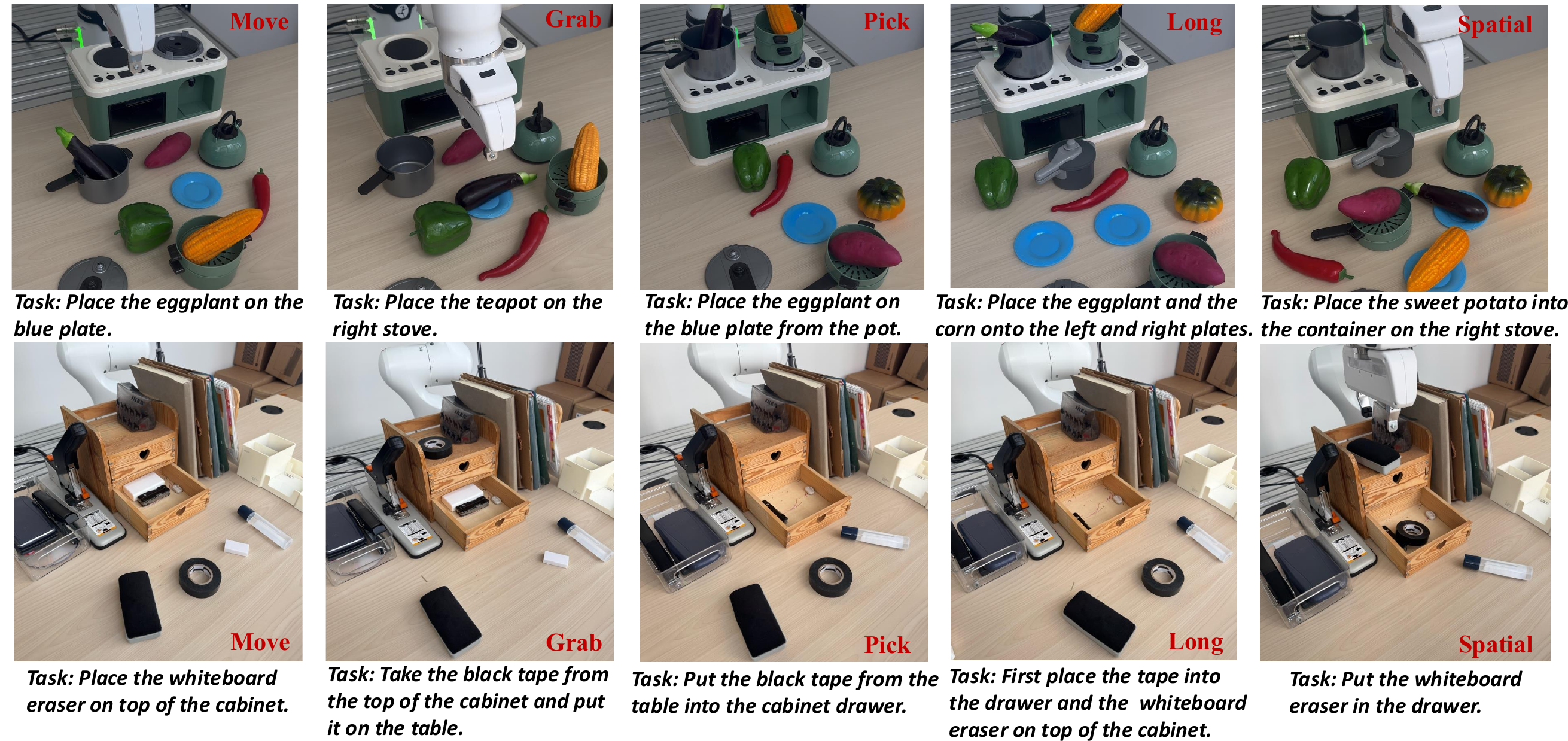}
    \vspace{-4mm}
    \caption{Representative manipulation tasks from the Kitchen (top) and Room (bottom) evaluation suites. The Kitchen tasks emphasize object selection under clutter, relative placement, and multi-object manipulation, whereas the Room tasks focus on object-centric spatial relations, precise stacking, and compositional long-horizon instructions. The language instruction associated with each task is shown below the corresponding scene.}
    \label{fig:realworld}
    \vspace{-2mm}
\end{figure*}

\input{Tables/realworld-table}

\vspace{-2mm}

\subsection{Layer-wise Geometric Probing}
\label{sec:layerwise_probe}

\paragraph{Geometric probing setup.}
We derive geometric labels from LIBERO observations using metric depth, camera calibration, and instance segmentation. Surface normals and log-depth gradients are computed from the reconstructed point map and pooled onto the $14\times14$ visual-token grid, while each task-relevant object center is obtained as the robust centroid of its segmented 3D points in the robot base frame. We freeze the complete VLA and extract image-token states
$\mathbf H_l\in\mathbb R^{B\times CN\times D}$
after every VLM decoder layer, additionally recording the representations immediately before and after each of the three \textbf{\textit{GA}} injections.

\paragraph{Layer-wise probes.}
For each layer, we train token-wise linear probes for surface normals and log-depth gradients:
\begin{equation} \small
    \hat{\mathbf n}_{l,p}
    =
    \operatorname{Norm}
    \left(
        \mathbf W^{n}_{l}\mathbf h_{l,p}+\mathbf b^{n}_{l}
    \right),
    \qquad
    \hat{\mathbf g}_{l,p}
    =
    \mathbf W^{g}_{l}\mathbf h_{l,p}+\mathbf b^{g}_{l},
\end{equation}
evaluated using mean cosine similarity and the coefficient of determination $R^{2}$, respectively. For object localization, token features are pooled using fractional instance mask $m_{o,p}$:
\begin{equation}
    \mathbf z_{l,o}
    =
    \frac{\sum_{p}m_{o,p}\mathbf h_{l,p}}
         {\sum_{p}m_{o,p}+\epsilon},
    \qquad
    \hat{\mathbf c}_{l,o}
    =
    \mathbf W^{c}_{l}\mathbf z_{l,o}+\mathbf b^{c}_{l},
\end{equation}
and performance is reported as the median Euclidean center error in centimeters. All probes use identical capacity and episode-disjoint splits, with frozen SigLIP and VGGT features included as semantic and geometric references.

\input{Tables/efficiency_cache}

\begin{figure}
    \centering
    \includegraphics[width=1\linewidth]{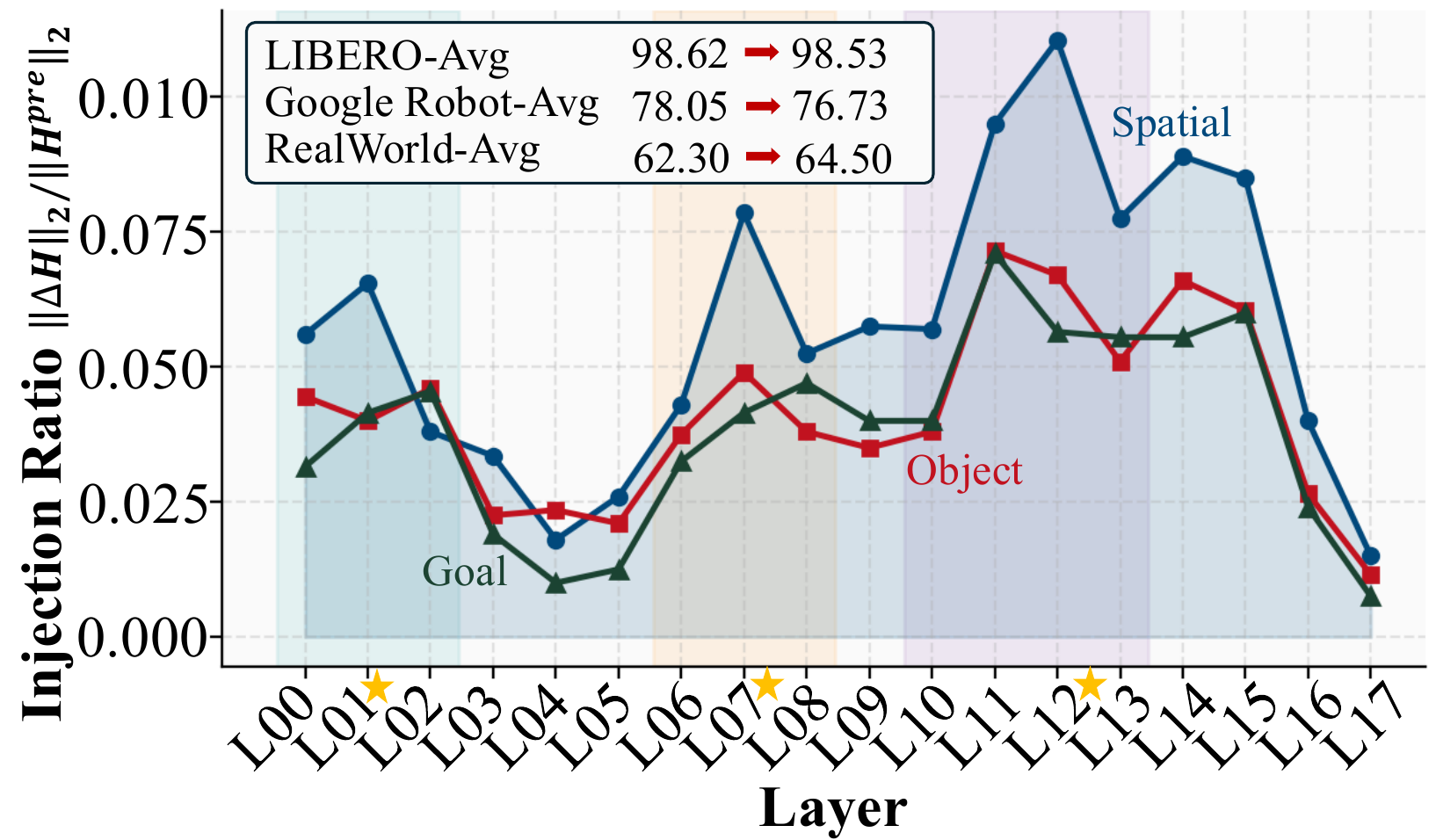}
    \vspace{-6mm}
    \caption{Selection of geometric injection layers. Reports the normalized \textbf{\textit{GA}} residual ratio $\lVert\Delta H_l\rVert_2/\lVert H_l^{\mathrm{pre}}\rVert_2$ obtained when adapters are attached to all backbone layers. The shaded regions reveal distinct shallow, middle, and deep response stages, while the markers denote the selected injection layers for $\pi_0$. The top-left box region shows the performance comparison of all-layer injection toward sparse injection.}
    \label{fig:injection}
    \vspace{-2mm}
\end{figure}

\paragraph{Results.}

\begin{figure}
    \centering
    \includegraphics[width=1\linewidth]{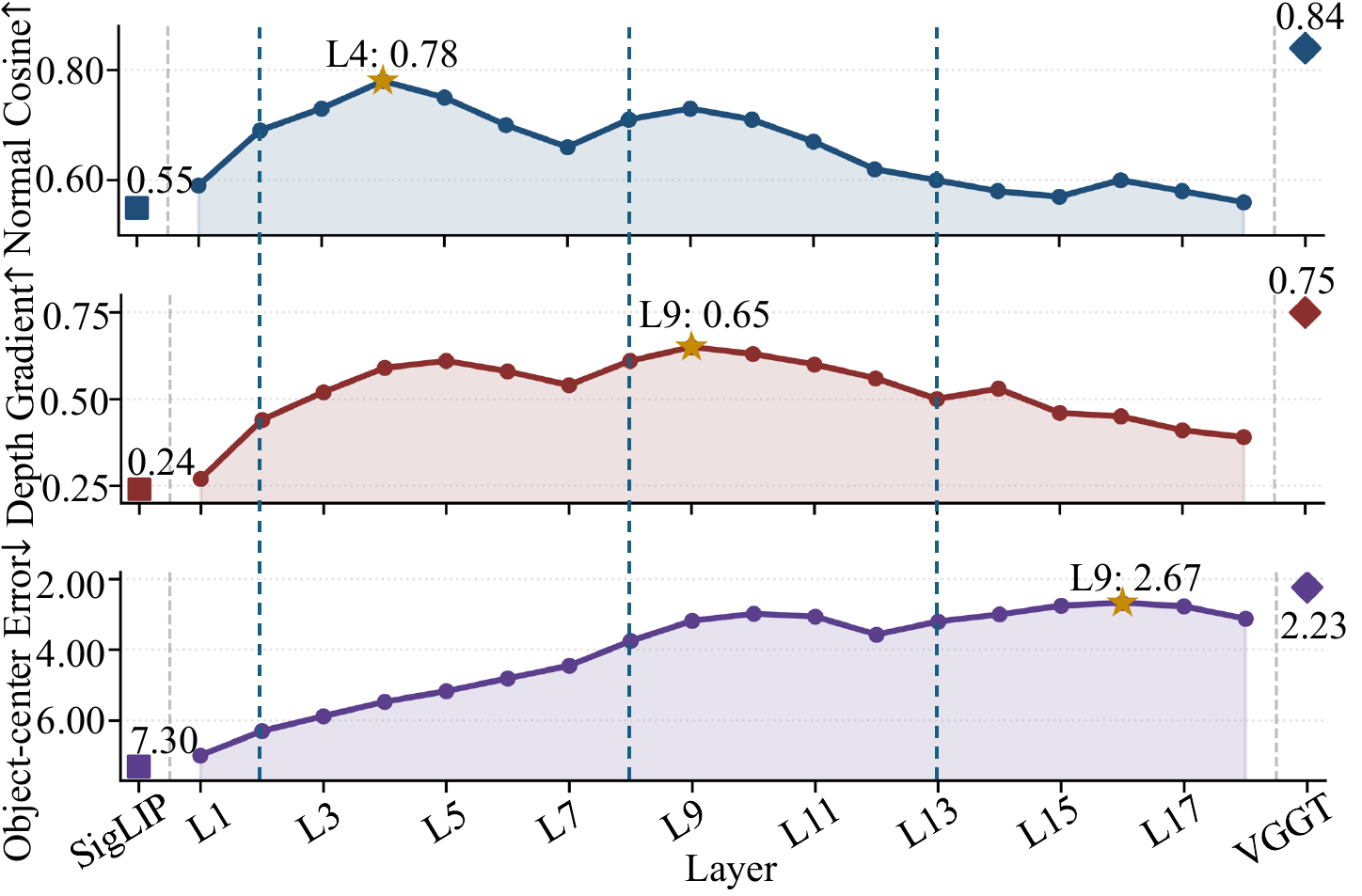}
    \caption{Layer-wise linear-probe analysis of geometric information encoded by \textbf{\textit{StrataVLA}}. Independent linear probes are evaluated on the frozen SigLIP features, all 18 VLM layers, and the raw VGGT representation. Vertical dashed lines indicate the three \textbf{\textit{GA}} injection layers.}
    \label{fig:probe}
    \vspace{-2mm}
\end{figure}

Figures~\ref{fig:injection} and~\ref{fig:probe} jointly reveal three distinct geometric response stages. The shallow response is concentrated around the first few blocks, the middle response rises again near Layers~7--8, and the strongest deep-stage response appears around Layers~11--13, particularly on LIBERO-Spatial. We therefore select Layers~2, 8, and 13 as representative injection points. Importantly, the probe maxima need not occur exactly at the injected block: the injected residual is subsequently transformed by later self-attention and feed-forward layers, so its most linearly decodable form may emerge one or two layers downstream. Surface-normal decoding peaks at Layer~4 with a cosine similarity of $0.78$, while depth-gradient decoding reaches $R^{2}=0.65$ at Layer~9. Object-center error decreases from $7.30$\,cm for SigLIP to $3.75$\,cm at Layer~8 and $2.98$\,cm at Layer~10, temporarily rises to $3.57$\,cm at Layer~12, and then recovers after the deep injection to $2.92$\,cm at Layer~13 and $2.64$\,cm at Layer~15. This late-stage recovery is consistent with the high injection magnitude observed in the deep response region of Fig.~\ref{fig:injection}.



\paragraph{Analysis.}
Experiments provide complementary evidence for layer selection. The injection-ratio curves identify where the policy actively requests geometric information, whereas the probes reveal what information remains linearly accessible after each request. Their agreement supports a staged organization: shallow injection strengthens local surface structure, middle injection consolidates depth and object localization, and deep injection restores metric object precision after semantic abstraction. The non-monotonic probe curves further suggest that intermediate reasoning can suppress geometric detail, making sparse updates near functional transitions more effective than dense injection. Finally, degradation after Layer~15 suggests that final backbone states increasingly prioritize action-conditioned semantics rather than preserving accurate general-purpose geometric representation.


\subsection{Inference Efficiency}

As shown in Table~\ref{tab:efficiency_cache}, the three sparse
\textbf{\textit{GA}} introduce only 4.8\,ms of additional latency,
corresponding to a 3.8\% overhead when the geometric feature is
already available. The cache hit rate varies with scene dynamics while LIBERO-Spatial obtains the lowest hit rate because changes in object layout and relative geometry more frequently invalidate the cached representation. Together with task-aware routing, the LRU cache suppresses 80.8\% of VGGT invocations on average and up to 88.0\% on LIBERO-Long, reducing the amortized VGGT cost from 120.0\,ms to 23.0\,ms on average and 14.4\,ms in the most reusable setting. Cache reuse has a negligible effect on task accuracy, improving the average success rate by only 0.04 points, which indicates that its primary benefit is computational efficiency rather than additional policy capacity. The cache contains at most eight entries per episode and requires only approximately 11.3\,MiB of GPU memory, while the \textit{\textbf{GAs}} are evaluated once during prefix construction and are not recomputed during the denoising steps.

\section{Conclusion}
We present \textbf{\textit{StrataVLA}}, a plug-and-play framework for hierarchical 3D grounding in VLA models. Layer-specific Geometry Adapters allow visual tokens at different backbone depths to retrieve task-relevant evidence from a shared geometric memory, yielding consistent gains across simulation and real-world tasks. Combined with task-aware routing and temporal reuse, \textbf{\textit{StrataVLA}} significantly reduces geometry-model invocations with negligible performance loss, demonstrating that hierarchical geometry injection is both effective and practical for spatially grounded robotic control.

\bibliography{aaai2027}

\appendix

\input{Appendix}

\end{document}

%% file: Tables/main-exp.tex
\begin{table*}[t]
  \caption{
    Success rates (\%) on the four standard LIBERO suites and the SimplerEnv Google Robot tasks. LIBERO results are reported as mean $\pm$ standard deviation where available, and published results are taken from their reported value. The reported \textbf{\textit{StrataVLA}} results are implemented on $\pi_0$ as the base policy; results based on OpenVLA-OFT are provided in the appendix. PCC, MN, and O/C denote Pick Coke Can, Move Near, and Open/Close Drawer, respectively.
}
  \label{tab:main_results}
  \centering
  \small
  \setlength{\tabcolsep}{6pt}
  \begin{tabular}{lcccclccc}
    \toprule
    \multirow{2}{*}{Method} & \multicolumn{4}{c}{Standard LIBERO Suites} & \multirow{2}{*}{Method} & \multicolumn{3}{c}{SimplerEnv Google}\\
    \cmidrule(r){2-5}
    \cmidrule(l){7-9}
    & Spatial & Object & Goal & Long-10 
    &
    & PCC & MN & O/C \\
    \midrule
    Octo~\cite{Team2024OctoAO} & 78.9$_{\pm \text{1.0}}$ & 85.7$_{\pm \text{1.2}}$ & 84.6$_{\pm \text{0.9}}$ & 50.9$_{\pm \text{1.2}}$ & Octo~\cite{Team2024OctoAO} & 17.0 & 4.2 & 22.7 \\

    OpenVLA~\cite{Kim2024OpenVLAAO} & 85.0$_{\pm \text{1.1}}$ & 88.6$_{\pm \text{0.9}}$ & 79.2$_{\pm \text{1.0}}$ & 53.6$_{\pm \text{1.0}}$ & OpenVLA~\cite{Kim2024OpenVLAAO} & 16.3 & 46.2 & 35.6 \\
    
    $\pi_0$~\cite{Black20240AV} & 96.8$_{\pm \text{1.3}}$ & 98.8$_{\pm \text{1.2}}$ & 95.8$_{\pm \text{1.5}}$ & 85.2$_{\pm \text{1.0}}$ & $\pi_0$~\cite{Black20240AV} & 88.0 & 80.3 & 56.0 \\

    SpatialVLA~\cite{Qu2025SpatialVLAES} & 88.2$_{\pm \text{0.5}}$ & 89.9$_{\pm \text{0.7}}$ & 78.6$_{\pm \text{0.6}}$ & 55.5$_{\pm \text{1.0}}$ & RT-1-X~\cite{brohan2023rt2visionlanguageactionmodelstransfer} & 56.7 & 31.7 & \textbf{59.7} \\
    
    GeoVLA~\cite{Sun2025GeoVLAE3} & 98.4 & 99.0 & 96.6 & \textbf{96.6} & RT-2-X~\cite{brohan2023rt2visionlanguageactionmodelstransfer} & 78.7 & 77.9 & 25.0 \\
    
    ForeAct~\cite{Zhang2026ForeActSY} & 97.3 & 99.8 & 97.3 & 95.4 & RoboVLM~\cite{Li2024WhatMI} & 77.3 & 61.7 & 43.5 \\
    
    $\pi_{0.5}$~\cite{Intelligence202505AV} & 97.3$_{\pm \text{1.7}}$ & 98.8$_{\pm \text{0.9}}$ & 96.9$_{\pm \text{1.2}}$ & 94.2$_{\pm \text{1.8}}$ & SpatialVLA~\cite{Qu2025SpatialVLAES} & 86.0 & 77.9 & 57.4 \\
    
    GeoPredict ~\cite{Qian2025GeoPredictLP} & 98.0$_{\pm \text{0.7}}$ & 98.2$_{\pm \text{0.7}}$ & 95.7$_{\pm \text{0.2}}$ & 94.0$_{\pm \text{1.0}}$ & FALCON~\cite{Zhang2025FromST} & 90.7 & 79.2 & 39.8 \\
    
    \midrule
    \rowcolor{gray!15} \textbf{\textit{StrataVLA} (Ours)} & \textbf{100} & \textbf{99.82}$_{\pm \text{0.2}}$ & \textbf{97.83}$_{\pm \text{1.0}}$ & 96.47$_{\pm \text{2.6}}$ & \textbf{\textit{StrataVLA} (Ours)} & \textbf{91.7} & \textbf{82.0} & 56.5 \\
    \bottomrule
  \end{tabular}
\end{table*}

%% file: Tables/realworld-table.tex
\begin{table*}[t]

\caption{\textbf{Evaluation in RealWorld scene.} Success rates (\%) on the in-house Kitchen and Room manipulation suites. Each policy is extensively fine-tuned using its native architecture, and results are reported as mean $\pm$ standard deviation over 20 independent evaluation trials. The five task groups evaluate object movement, grasping, object picking, long-horizon execution, and spatial-relation reasoning, respectively.}
\centering
\small
\setlength{\tabcolsep}{5pt}

\begin{tabular}{lcccccccccc}
\toprule

\multirow{2}{*}{Method}
& \multicolumn{5}{c}{Kitchen}
& \multicolumn{5}{c}{Room}
\\

\cmidrule(lr){2-6}
\cmidrule(lr){7-11}

& Move & Grab & Pick & Long & Spatial
& Move & Grab & Pick & Long & Spatial 
\\

\midrule

Octo~\cite{Team2024OctoAO}
& 20$_{\pm \text{10}}$ & 10$_{\pm \text{5}}$ & 25$_{\pm \text{5}}$ & 0 & 15$_{\pm \text{10}}$
& 10$_{\pm \text{5}}$ & 0 & 20$_{\pm \text{10}}$ & 0 & 0 \\

OpenVLA~\cite{Kim2024OpenVLAAO}
& 35$_{\pm \text{15}}$ & 30$_{\pm \text{10}}$ & 45$_{\pm \text{5}}$ & 0 & 45$_{\pm \text{10}}$
& 30$_{\pm \text{10}}$ & 10$_{\pm \text{5}}$ & 30$_{\pm \text{10}}$ & 0 & 0\\

$\pi_0$~\cite{Black20240AV}
& 50$_{\pm \text{5}}$ & 35$_{\pm \text{5}}$ & 55$_{\pm \text{15}}$ & 15$_{\pm \text{5}}$ & 75$_{\pm \text{5}}$
& 40$_{\pm \text{5}}$ & 20$_{\pm \text{10}}$ & 50$_{\pm \text{5}}$ & 10$_{\pm \text{10}}$ & 15$_{\pm \text{5}}$ \\

$\pi_{0.5}$~\cite{Intelligence202505AV}
& 65$_{\pm \text{5}}$ & 35$_{\pm \text{10}}$ & 60$_{\pm \text{10}}$ & 30$_{\pm \text{10}}$ & 75$_{\pm \text{5}}$ 
& 45$_{\pm \text{10}}$ & 25$_{\pm \text{15}}$ & 65$_{\pm \text{5}}$ & 25$_{\pm \text{10}}$ & 15$_{\pm \text{10}}$ \\

\midrule
\rowcolor{gray!15}

\textbf{\textit{StrataVLA (Ours)}}
& \textbf{80}$_{\pm \text{10}}$ & \textbf{65}$_{\pm \text{5}}$ & \textbf{80}$_{\pm \text{5}}$ & \textbf{40}$_{\pm \text{5}}$ & \underline{75}$_{\pm \text{10}}$
& \textbf{60}$_{\pm \text{5}}$ & \textbf{40}$_{\pm \text{5}}$ & \textbf{70}$_{\pm \text{10}}$ & \textbf{35}$_{\pm \text{15}}$ & \textbf{50}$_{\pm \text{15}}$ \\

\bottomrule
\end{tabular}

\label{tab:realworld}
\end{table*}

%% file: Tables/efficiency_cache.tex
\begin{table*}[t]
\centering
\caption{
Estimated inference latency and LRU-cache statistics of StrataVLA
on an NVIDIA RTX 4090 with batch size one. The three-GA setting
isolates adapter overhead with precomputed geometric memory.
Cache hit and write rates are measured over geometry-active steps,
whereas the skip rate additionally includes router bypasses.
}
\label{tab:efficiency_cache}
\small
\setlength{\tabcolsep}{3.0pt}
\renewcommand{\arraystretch}{0.96}

\begin{tabular}{
lrrrr
@{\hspace{9pt}}
lrrrrrr
}
\toprule
\multicolumn{5}{c}{\textbf{Per-action-chunk inference latency}}
&
\multicolumn{7}{c}{\textbf{Cache statistics and success on LIBERO}}
\\
\cmidrule(r){1-5}
\cmidrule(l){6-12}

Setting
& GA
& VGGT
& Total
& $\Delta$
&
Suite
& Hit
& Write
& Skip
& Peak
& $\mathrm{SR}_{-C}$
& $\mathrm{SR}_{+C}$
\\

& \multicolumn{3}{c}{Latency (ms)}
& vs.\ $\pi_0$
&
& \multicolumn{3}{c}{Rate (\%)}
& Entries
& \multicolumn{2}{c}{Success (\%)}
\\
\midrule

Base $\pi_0$
& -- & -- & 125.0 & --
&
Spatial
& 68.2 & 31.8 & 70.1 & 8 & 100.00 & 100.00
\\

3-GA only
& 4.8 & -- & 129.8 & $+3.8\%$
&
Object
& 76.4 & 23.6 & 79.0 & 8 & 99.82 & 99.90
\\

Cache miss
& 4.8 & 120.0 & 250.2 & $+100.2\%$
&
Goal
& 82.6 & 17.4 & 86.1 & 6 & 97.83 & 97.70
\\

Cache hit
& 4.8 & -- & 130.9 & $+4.7\%$
&
Long
& 85.4 & 14.6 & 88.0 & 7 & 96.31 & 96.66
\\

Amortized
& 4.8 & 23.0 & 153.8 & $+23.0\%$
&
Average
& 78.2 & 21.8 & 80.8 & 7.3 & 98.53 & 98.57
\\

\bottomrule
\end{tabular}
\vspace{-2mm}
\end{table*}

%% file: Appendix.tex
\section{Additional Training and Data Details}
\label{app:training_details}

This appendix provides the complete data mixture, sampling,
optimization, and training protocols used in our experiments.

\begin{table*}[t]
\centering
\caption{
Two-stage optimization settings. ``Backbone'' denotes
SigLIP--PaliGemma for $\pi_0$ and the Prismatic VLM for
OpenVLA-OFT. ``Action'' denotes the flow-matching Action Expert for
$\pi_0$ and the continuous parallel-decoding head for OpenVLA-OFT.
A dash indicates that the corresponding module is frozen.
}
\label{tab:optimization_details}
\begin{tabular}{llrrrrrrr}
\toprule
Dataset / Base
& Stage
& Steps
& Batch
& Backbone LR
& Action LR
& GA LR
& Router LR
& Warm-up \\
\midrule
LIBERO / $\pi_0$
& I
& 8K
& 128
& --
& --
& $1.0{\times}10^{-4}$
& $3.0{\times}10^{-4}$
& 500 \\
&
II
& 72K
& 128
& $2.0{\times}10^{-5}$
& $5.0{\times}10^{-5}$
& $8.0{\times}10^{-5}$
& $1.0{\times}10^{-4}$
& 1,000 \\
\midrule
SimplerEnv / $\pi_0$
& I
& 7K
& 128
& --
& --
& $1.0{\times}10^{-4}$
& $3.0{\times}10^{-4}$
& 500 \\
&
II
& 63K
& 128
& $2.0{\times}10^{-5}$
& $5.0{\times}10^{-5}$
& $8.0{\times}10^{-5}$
& $1.0{\times}10^{-4}$
& 1,000 \\
\midrule
Real world / $\pi_0$
& I
& 3K
& 64
& --
& --
& $1.0{\times}10^{-4}$
& $2.0{\times}10^{-4}$
& 300 \\
&
II
& 27K
& 64
& $1.0{\times}10^{-5}$
& $3.0{\times}10^{-5}$
& $5.0{\times}10^{-5}$
& $1.0{\times}10^{-4}$
& 500 \\
\midrule
LIBERO / OpenVLA-OFT
& I
& 8K
& 128
& --
& --
& $1.0{\times}10^{-4}$
& $3.0{\times}10^{-4}$
& 500 \\
&
II
& 72K
& 128
& $1.0{\times}10^{-5}$
& $3.0{\times}10^{-5}$
& $5.0{\times}10^{-5}$
& $1.0{\times}10^{-4}$
& 1,000 \\
\bottomrule
\end{tabular}
\end{table*}

\subsection{Dataset Construction and Sampling}
\label{app:data_sampling}

\paragraph{LIBERO mixture.}
We construct a single training set from all four standard LIBERO
suites, containing 40 tasks and 2,000 expert demonstrations in total.
Rather than uniformly sampling the four suites, we use the
difficulty-aware mixture
\begin{equation*}
p_{\mathrm{suite}}
=
\left[
0.30,\,
0.20,\,
0.20,\,
0.30
\right]
\label{eq:libero_sampling}
\end{equation*}
for LIBERO-Spatial, LIBERO-Object, LIBERO-Goal, and LIBERO-Long,
respectively. The larger weights assigned to Spatial and Long reflect
their stronger requirements for metric spatial reasoning and
long-horizon geometric consistency.

Sampling is hierarchical. We first sample a suite according to
the former equation, then uniformly sample one of its ten
tasks, one demonstration from that task, and a valid anchor timestep
from the selected demonstration. This episode-first procedure prevents
the longer trajectories in LIBERO-Long from being implicitly
over-represented merely because they contain more transitions. The
same mixed policy is evaluated separately on the four suites.

\paragraph{SimplerEnv--Google Robot.}
For the Google Robot experiments, we fine-tune the pretrained policy
on the complete training split of the RT-1 Fractal dataset. We use only real-robot trajectories from Fractal and do not train on observations, actions, or rollouts generated by SimplerEnv. Consequently, SimplerEnv is used strictly as
a real-to-sim evaluation environment.

To reduce the dominance of frequent low-level behaviors in Fractal,
we group episodes by their normalized language instruction and sample
an instruction group $c$ using temperature sampling,
\begin{equation*}
p(c)
=
\frac{n_c^{\,0.5}}
{\sum_{c'} n_{c'}^{\,0.5}},
\label{eq:fractal_sampling}
\end{equation*}
where $n_c$ is the number of episodes associated with instruction
$c$. An episode and a valid action window are then sampled uniformly
from the selected group. We reserve 2\% of the episodes as a
development set using an episode-disjoint split.

The Fractal end-effector deltas and gripper commands are converted to
the native action representation of the base policy. Each continuous
action dimension is clipped at its training-set 1st and 99th
percentiles and normalized independently. Evaluation follows the
Google Robot Visual Matching protocol on Pick Coke Can, Move Near,
and Open/Close Drawer. No benchmark-specific simulation fine-tuning
or online adaptation is performed.

\paragraph{Real-world data.}
Our real-world benchmark contains five semantic task categories
(\emph{Move}, \emph{Grab}, \emph{Pick}, \emph{Long}, and
\emph{Spatial}) instantiated in both the Kitchen and Room scenes.
For each category, we collect 80 teleoperated demonstrations in total,
with 40 demonstrations collected in each scene. The resulting dataset
therefore contains
\begin{equation*}
5\ \text{categories}
\times
2\ \text{scenes}
\times
40\ \text{episodes}
=
400\ \text{episodes}.
\end{equation*}
The average trajectory length is approximately 280 control steps.

The split is performed at the episode level and stratified over all
ten category--scene pairs. For every pair, 36 episodes are used for
training and four for development, yielding 360 training episodes and
40 development episodes. Training batches are constructed by first
sampling one of the five task categories uniformly, then sampling one
of the two scenes uniformly, followed by an episode and a valid
timestep. This avoids over-sampling the longer demonstrations in the
Long category and gives equal exposure to the two visual domains.

\subsection{Two-Stage Optimization}
\label{app:two_stage_training}

All experiments use AdamW with
$(\beta_1,\beta_2)=(0.9,0.95)$, $\epsilon=10^{-8}$, weight decay
$0.01$, and gradient-norm clipping at $1.0$. Biases, LayerNorm
parameters, and scalar gates are excluded from weight decay. Learning
rates are linearly warmed up and then cosine-decayed to 10\% of their
peak values. We reset the optimizer and learning-rate scheduler when
transitioning from Stage~I to Stage~II.

In Stage~I, the pretrained policy is frozen and only the GAs and
dual-modal routing heads are optimized. This allows the
zero-initialized geometric residual paths to become informative
without perturbing the pretrained policy. In Stage~II, all policy
components are jointly optimized, while VGGT remains frozen. For
$\pi_0$, the trainable policy components include SigLIP, PaliGemma,
the Action Expert, the action-output projection, the GAs, and the
router. For OpenVLA-OFT, Stage~II includes the fused visual encoder,
visual projector, Llama decoder, continuous action head, GAs, and
router. The effective batch sizes and group-wise peak learning rates are
summarized in Table~\ref{tab:optimization_details}.

For LIBERO and SimplerEnv, the total optimization budgets are 80K and
70K updates, respectively. For the smaller real-world dataset, we use
30K updates together with lower Stage~II learning rates to reduce
overfitting to the ten observed task--scene combinations. The
development split is used only for checkpoint selection, and the
checkpoint with the lowest development action loss within the final
5K updates is retained.

For $\pi_0$, we preserve the original flow-matching objective and use
ten Euler integration steps at inference. OpenVLA-OFT retains parallel
action decoding, continuous action prediction, action chunking, and
its native $\ell_1$ regression objective. We do not introduce an
additional action loss for either backbone.

\begin{table*}[t]
\centering
\caption{
Success rates (\%) of OpenVLA-OFT and its StrataVLA variant on the
four LIBERO suites. Avg. is the unweighted average over suites.
}
\label{tab:openvla_appendix}
\setlength{\tabcolsep}{4.5pt}
\begin{tabular}{lccccc}
\toprule
Method
& Spatial
& Object
& Goal
& Long
& Avg. \\
\midrule
OpenVLA-OFT
& 96.2
& 98.3
& 96.2
& 90.7
& 95.35 \\
\textbf{StrataVLA (OpenVLA-OFT)}
& \textbf{99.2}
& \textbf{99.4}
& \textbf{97.4}
& \textbf{95.6}
& \textbf{97.90} \\
\bottomrule
\end{tabular}
\end{table*}

\subsection{Router Warm-up and Weak Supervision}
\label{app:router_training}

The zero-initialized GA output projection initially produces an
exactly zero residual. To ensure that the adapters receive a stable
optimization signal before routing becomes selective, we force the
geometry coefficient to one during the first $N_{\mathrm{force}}$
updates:
\begin{equation}
N_{\mathrm{force}}
=
\begin{cases}
500, & \text{LIBERO and SimplerEnv},\\
300, & \text{real-world training}.
\end{cases}
\end{equation}
During this interval, the router is still optimized by its auxiliary
loss, but its prediction is not used to scale the GA residual. After
the forced-routing warm-up, the continuous coefficient from the
dual-modal router is used end-to-end.

The router supervision combines an instruction prior with an offline
task-level gain estimate. An instruction receives a positive geometry
label when it explicitly contains a metric or relational concept,
including relative directions, containment, stacking, occlusion,
between-object placement, or depth-dependent selection. After
Stage~I, we additionally estimate the benefit of geometry for each
task using the relative reduction in development action loss:
\begin{equation}
g_k
=
\frac{
\mathcal{L}^{\mathrm{base}}_k
-
\mathcal{L}^{\mathrm{geo}}_k
}{
\mathcal{L}^{\mathrm{base}}_k+\epsilon
},
\label{eq:offline_gain}
\end{equation}
where $\mathcal{L}^{\mathrm{base}}_k$ and
$\mathcal{L}^{\mathrm{geo}}_k$ are obtained with the geometry path
disabled and enabled, respectively, on task $k$.

We assign $y_{\mathrm{gate}}=1$ when either the instruction prior is
positive or $g_k\geq0.02$. A sample is assigned
$y_{\mathrm{gate}}=0$ when the instruction prior is negative and
$g_k\leq0.005$. Samples between these thresholds are treated as
unlabeled and contribute only to the action objective. Low-quality
frames satisfying $q_t<\tau_q$ are also excluded from the auxiliary
router loss because their geometry branch is deterministically
disabled by the image-quality constraint.

For labeled samples, the complete training objective is
\begin{equation}
\mathcal{L}
=
\mathcal{L}_{\mathrm{act}}
+
0.05\,
\mathcal{L}_{\mathrm{gate}},
\qquad
\mathcal{L}_{\mathrm{gate}}
=
\operatorname{BCE}
\left(
\alpha_t,
y_{\mathrm{gate}}
\right),
\label{eq:appendix_training_loss}
\end{equation}
where $\mathcal{L}_{\mathrm{act}}$ is the flow-matching regression
loss for $\pi_0$ or the continuous $\ell_1$ loss for OpenVLA-OFT.

We use the image-quality threshold $\tau_q=0.25$, initialize
$w_{\mathrm{text}}=w_{\mathrm{vis}}=1.5$ and $b=1.0$, and apply a
dropout rate of $0.1$ inside both routing heads. During training, VGGT
is evaluated for all frames that pass the image-quality constraint,
and the continuous $\alpha_t$ scales the injected residual. We do not
apply the discrete routing bypass during optimization, which preserves
gradient flow through the router. At inference, coefficients below
$\tau_{\mathrm{route}}=0.10$ are set to zero, bypassing both VGGT and
all GAs. The LRU cache is enabled only at inference and is cleared at
the beginning of every episode.

\section{Additional OpenVLA-OFT Results}
\label{app:openvla_results}

To evaluate whether hierarchical geometric grounding transfers beyond
the $\pi_0$ architecture, we additionally instantiate StrataVLA on
OpenVLA-OFT. We preserve the original fused SigLIP--DINOv2 encoder,
visual projector, Llama decoder, parallel continuous action head, and
$\ell_1$ training objective. Three independent GAs are inserted after
decoder layers $\{4,15,24\}$ using zero-based indexing. The geometry
expert and geometric-memory extraction procedure are identical to the
$\pi_0$ implementation.

Both models in Table~\ref{tab:openvla_appendix} use the same LIBERO
data mixture and optimization budget described in
Appendix~\ref{app:training_details}. StrataVLA consistently improves
the OpenVLA-OFT backbone, with the largest gains on LIBERO-Spatial and
LIBERO-Long, where metric relations and long-horizon geometric
consistency are most important.

Relative to OpenVLA-OFT, hierarchical geometric injection improves
the average success rate by 2.55 percentage points. These results indicate that StrataVLA is not specific to the joint-attention or flow-matching formulation of
$\pi_0$, but also transfers effectively to a decoder-only VLA with a
continuous regression action head.